\documentclass[10pt,twocolumn,letterpaper]{article}

\usepackage[pagenumbers]{cvpr} 
\usepackage{multirow}

\usepackage{arydshln}
\usepackage[dvipsnames]{xcolor}

\newcommand{\sysname}{SMoLA\xspace}
\newcommand{\pali}[2]{PaLI-#1$_\texttt{#2}$}
\newcommand{\sysversion}[4][O]{SMoLA$_{\texttt{#1}}^{\texttt{#2}}$-\pali{#3}{#4}}

\definecolor{cvprblue}{rgb}{0.21,0.49,0.74}
\usepackage[pagebackref,breaklinks,colorlinks,citecolor=cvprblue]{hyperref}

\def\paperID{12801} 
\def\confName{CVPR}
\def\confYear{2024}

\title{Omni-SMoLA: Boosting Generalist Multimodal Models with Soft Mixture of Low-rank Experts}

\author{First Author\\
Affiliation / Address line 1\\
Affiliation / Address line 2\\
{\tt\small firstauthor@example.com}
\and
Second Author\\
Different Affiliation / Address line 1\\
Different Affiliation / Address line 2\\
{\tt\small secondauthor@example.com}
}

\author{Jialin Wu \qquad  Xia Hu  \qquad Yaqing Wang \qquad Bo Pang \qquad Radu Soricut \\
Google Research\\
{\tt\small \{jialinwu, xiahu, yaqingwang, bopang, rsoricut\}@google.com}
}

\begin{document}
\maketitle
\begin{abstract}
Large multi-modal models (LMMs) exhibit remarkable performance across numerous tasks. However, generalist LMMs often suffer from performance degradation when tuned over a large collection of tasks. Recent research suggests that Mixture of Experts (MoE) architectures are useful for instruction tuning, but for LMMs of parameter size around O(50-100B), the prohibitive cost of replicating and storing the expert models severely limits the number of experts we can use.
We propose Omni-\sysname, an architecture that uses the Soft MoE approach to (softly) mix many multimodal low rank experts, and avoids introducing a significant number of new parameters compared to conventional MoE models. The core intuition here is that the large model provides a foundational backbone, while different lightweight experts residually learn specialized knowledge, either per-modality or multimodally. Extensive experiments demonstrate that the \sysname approach helps improve the generalist performance across a broad range of generative vision-and-language tasks, achieving  new SoTA generalist performance that often matches or outperforms single specialized LMM baselines, as well as new SoTA specialist performance.
\end{abstract}    
\section{Introduction}
\label{sec:intro}
Large multimodal models (LMMs) \cite{tsimpoukelli2021multimodal,chen2023pali,chen2023palix,chen2023pali5b,li2023blip2,driess2023palme} demonstrate remarkable performance on a variety of tasks including visual question answering, image captioning, visual document understanding, etc.
To date, the best performance on most of these tasks is achieved by so-called {\it specialist} LMMs, but their large scale makes it impractical to deploy a multitude of such specialists at once.
As a result, so-called {\it generalist} LMMs emerge as an obvious choice, where such a model is trained and deployed to handle a wide range of tasks using the same set of model parameters. 

Building  a  single generalist  model to  solve  multiple tasks remains challenging. 
A straightforward approach is to fine-tune the model parameters with supervised data representing multiple tasks. However, recent research suggests that it causes non-negligible performance degradation compared to the performance of a single-task specialist \cite{chen2023palix}. It is likely that, even though these tasks share the same configuration of modalities (e.g., image + text as input, text as output), what the model needs to solve for is significantly diverse -- for instance, some tasks require recognizing the fine-grained identity of visual content, others may rely on world-knowledge outside of the visual scene, while others require reading and understanding texts from images. 

Recent work \cite{shen2023mixtureofexperts} show that Mixture-of-Experts (MoE) models stand to benefit more from instruction tuning compared to dense models, and serve as good candidate architectures for building generalist large language models. Intuitively, this should work well because different expert modules can specialize and handle different tasks. However, there is an obvious issue with applying the MoE design on Transformer blocks for large-scale models: the different transformer blocks result in replicating the model parameters using high-rank experts.
This creates a situation in which the scale of each expert model block compared to their dense-model counterparts is much more limited.

In this work, we address the aforementioned limitations by introducing Omni-\sysname, an architecture that efficiently mixes many multi-modal low rank experts.
Using this architecture, we demonstrate strong capabilities for adapting pretrained models to tackle specialized tasks. The core intuition is that a large pretrained (or instruction-tuned) model provides a foundational backbone of capabilities (we denote this model by $\theta^*$), while different lightweight experts learn additional specializations (which can be knowledge, style, or capabilities).
In particular, for the modalities considered in this paper (text \& vision), the Omni-\sysname architecture consists of three sets of experts, focusing on text tokens, visual tokens and multimodal tokens, respectively, in order to satisfy different needs from various tasks.

In general, the \sysname design has several important proprieties.
First, due to its adoption of the low rank expert design \cite{softmoe} and unlike conventional MoE transformer models \cite{Lepikhin2020GShardSG,Fedus2021SwitchTS,du2022glam,shen2023mixtureofexperts}, the total parameter count is not proportional to the product of expert counts and the parameter counts in each expert as the backbone still contains majority of the parameters. This allows it to bypass the limitation on the number of experts used, which helps achieve better generalist performance.
Second, this design is potentially compatible with any large model architecture, either dense or MoE.
And, last but not least, it allows for the freedom to potentially adopt different model architectures between the pretraining stage and multi-task learning (or instruction tuning) stage. 

We evaluate the Omni-\sysname approach on a variety of settings, starting from PaLI-3 \cite{chen2023pali5b} (a 5B LMM) and PaLI-X \cite{chen2023palix} (a 55B LMM), models that have current state-of-the-art (SOTA) performance across a wide range of vision-language benchmarks.
The settings include various image captioning tasks and visual question answering tasks, and we experiment with possible combinations in terms of model specialization.
We find that:
(1) Omni-\sysname achieves better average performance compared to full-model fine-tuning baselines for both PaLI-3 and PaLI-X; our experiments show that it achieves new SoTA results on multiple vision-language benchmarks, both under generalist settings and under specialist settings;
(2) the performance improves with the introduction of the Omni experts, and also increases with the number of experts;
(3) in spite of the added modules and a large number of experts per module, the inference speed is only slightly slower compared the base models, indicating the efficiency of this design.

\section{Related Work}
\subsection{Large Multi-modal Models}
Inspired by the success of Large Language Model~\cite{brown2020language,BERT,cohen2022lamda}, there is a growing interest of building large multi-modal models~(LMMs)~\cite{chen2023pali,chen2023pali5b,driess2023palme,li2023blip2} that is designed to understand both vision and language signals simultaneously~\cite{li2023large,driess2023palme}.
The main approach is to integrate a pretrained image encoder, which represents images as a sequence of continuous embeddings, to autoregressive language model~\cite{tsimpoukelli2021multimodal,chen2023pali,chen2023palix,chen2023pali5b,li2023blip2,driess2023palme}. 
PaLI series of works~\cite{chen2023pali,chen2023palix,chen2023pali5b} integrate pretrained ViT models~\cite{dosovitskiy2021an} to encoder-decoder language framework.
PaLM-E~\cite{driess2023palme} incorporate vision encoders as sensor modalities to language model and enable the model to process multiple images in text sentences in a flexible way. 
BLIP-2~\cite{li2023blip2} propose a lightwieght Querying Transformer to leverage frozen pre-trained image encoder and language model for multimodal tasks. 

\subsection{Parameter-Efficient Fine-Tuning}
Recently the success of scaling up model size encourage the development of larger language models~\cite{chen2023palix,treviso2023efficient,chowdhery2022palm,rae2021scaling}. 
Meanwhile, parameter-efficient fine-tuning~\cite{treviso2023efficient,houlsby2019parameter,bapna2019simple,rebuffi2017learning,pfeiffer2020adapterhub,hu2022lora,sung2021training} aims to discover a more efficient solution to adapt large models to particular downstream tasks.
Instead of full model fine-tuning which updates the entire set of model parameters, parameter-efficient fine-tuning updates or adds a relatively small number of parameters and leaves the rest of model parameters fixed~\cite{treviso2023efficient}. 
FISH Mask\cite{sung2021training} applies a fixes sparse mask on model parameters and only updates mask-selected parameters.
Adapters~\cite{houlsby2019parameter,bapna2019simple,rebuffi2017learning,pfeiffer2020adapterhub} inserts new trainable dense layers into Transformer and leave the original model parameters frozen. 
Prefix-tuning~\cite{li2021prefix} and prompt-tuning~\cite{lester2021power} freeze parameters of the model and learn continuous prompts. 
LoRA~\cite{hu2022lora} injects trainable low-rank decomposition matrices into every layer of Transformer and freezes the pretrained language model parameters. 
In particular, LoRA shows outstanding capability to achieve competitive or even better performance than fine-tuning with only $0.1\%$ trainable parameters~\cite{hu2022lora,wang2023non}

\subsection{Mixture-of-Experts for Multitask Learning}
Mixture-of-Experts (MoE) architectures are centered around enhancing conditional computation capabilities and scale parameters in neural architectures such as Transformers. The MoE transformer models~\cite{shazeer2017outrageously, fedus2021switch,lepikhin2020gshard,zuo2021taming} typically employ $N$ feed-forward networks, referred to as ``experts''.
Each of these experts has its unique set of trainable weights, enabling them to craft distinct representations for each input token based on contextual information.
Multitask learning~(MTL), a popular ML topic for many years, aims at finding solutions to simultaneously improving performance on multiple tasks of interests \cite{caruana1997multitask,lin2019pareto}.
Recently, mixture-of-experts~(MoE)~\cite{Jacobs1991AdaptiveMO, Jordan1993HierarchicalMO,Shazeer2017OutrageouslyLN} approaches have become a promising approach for MTL~\cite{fan2022m3vit}, benefiting from its strategy of separating the parameter space and adopt relevant model part to different tasks. 

Inspired by these advances, there is an increasing interest in investigating the application of MoEs in Transformer-based large models.
Some methods adopt MoE in Transformer structure of large language models~\cite{Lepikhin2020GShardSG,Fedus2021SwitchTS,du2022glam,shen2023mixtureofexperts}. 
Gshard~\cite{Lepikhin2020GShardSG} introduces the idea of scaling Transformer in LMMs with MoE layers, in which the feed forward layer of every other Transformer is replaced by a Sparsely-Gated MoE layer.
This MoE Transformer structure is then used in ~\cite{du2022glam} to develop a family of Decoder-only language models, and ~\cite{shen2023mixtureofexperts} which finds MoE modified LLM models benefits more from instruction tuning than dense LLMs.

Some other methods explore combining MoE with parameter-efficient fine-tuning.
AdaMix~\cite{wang2022adamix} proposes a mixture-of-adapters mechanism to improve per-task tuning performance.
The most relevant work is the concurrent research \cite{zadouri2023pushing} that introduces mixture of LoRA by weighted summing of different LoRA outputs. While conceptually similar, our \sysname approach differs by having significantly lower computational cost, and also allowing hundreds of experts to handle single and multiple modalities with negligible inference speed cost.
We find that scaling to hundreds of experts is crucial to attaining improved generalist performance.

\begin{figure*}[t!]
  
 		\centering
	\includegraphics[clip, trim=0cm 5cm 5cm 1cm, width=1.4\columnwidth]{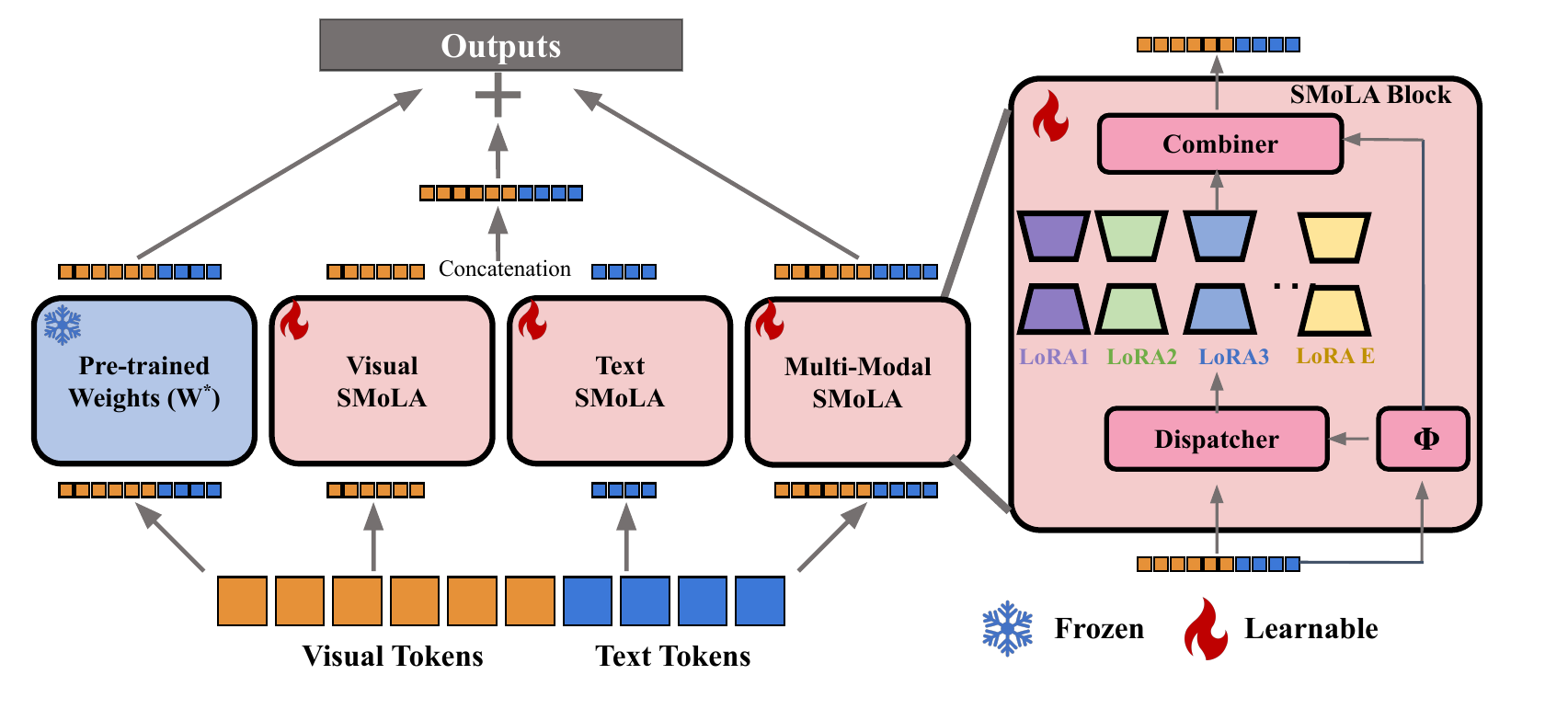}
 	  	\caption{Omini-\sysname model architecture contains three \sysname blocks that take as input visual tokens, text tokens and multimodal tokens. Each such block employs a large set of low-rank experts.}
 	 	 	\label{fig:framework}
 
\end{figure*}
\section{Methodology}

\subsection{Preliminaries}

\paragraph{Low-rank Adaptation (LoRA).}
Low-Rank Adaptation (LoRA)~\cite{hu2021lora} is a technique designed to enhance the adaptability of pretrained transformer models to new tasks with a minor increase in trainable parameter counts. It can be applied on any linear layers, offering great compatibility with recent large models. 

We denote $\texttt{W} \in \texttt{R}^{\texttt{d}_1 \times \texttt{d}_2}$ as the weight matrix for a linear layer from the large model. LoRA introduces two low-rank matrices  $\texttt{W}^{\texttt{in}} \in \texttt{R}^{\texttt{r} \times \texttt{d}_1}$ and $\texttt{W}^{\texttt{out}} \in \texttt{R}^{\texttt{d}_2 \times \texttt{r}}$ for each layer, where $\texttt{r} \ll \texttt{min}\{\texttt{d}_1, \texttt{d}_2\}$. The $\texttt{W}^{\texttt{in}}$ and $\texttt{W}^{\texttt{out}}$ are consecutively applied to the input of the linear layer to project the input to a low rank space and then project back to the output space. The adapted weights $\texttt{W}'$ can be represented as $
\texttt{W}'= \texttt{W}  + \texttt{W}^{\texttt{out}}\texttt{W}^{\texttt{in}}$.
As the rank of $\texttt{W}^{\texttt{in}}$ and $\texttt{W}^{\texttt{out}}$ is limited by $\texttt{r}$ and typically much smaller than $\texttt{d}_1$ and $\texttt{d}_2$, the LoRA approach serves as a compact and efficient adaptation mechanism.
\paragraph{Soft Mixture of Experts (Soft MoE).}
We briefly recap the Soft MoE model in this section (details can be found in \cite{softmoe}).
The core idea is to learn a dispatcher module that can dispatch input tokens to different experts, and a combiner module that can combine the results from all the experts and project them back to the original token space. 

We denote the input to the transformer block as $\textbf{X} \in \texttt{R}^{\texttt{N} \times \texttt{d}_1}$, consisting of $\texttt{N}$ tokens. Soft MoE introduces a routing matrix $\Phi \in \texttt{R}^{\texttt{E} \times \texttt{d}_1}$ that corresponds to $\texttt{E}$ experts. The dispatcher and combiner are represented by Eq. \ref{eq:dispatcher} and \ref{eq:combiner}: \texttt{norm} denotes l2 normalization and $\alpha$ is a learnable scalar.
\begin{align}
    \textbf{D} = \texttt{softmax}(\alpha \cdot \texttt{norm}(\Phi) \texttt{norm}(\textbf{X})^T, \texttt{axis} \texttt{=} \texttt{1}) \label{eq:dispatcher}\\
\textbf{C} = \texttt{softmax}(\alpha \cdot \texttt{norm}(\Phi) \texttt{norm}(\textbf{X})^T, \texttt{axis} \texttt{=} \texttt{0}) \label{eq:combiner}
\end{align}

Each expert model $f_\texttt{i}$ (usually MLP Blocks) operates on the corresponding slice of dispatched inputs $\tilde{x}_i = (\textbf{D}\textbf{X})_{\texttt{i}, :}$ to produce   $\tilde{y}_\texttt{i} = f_\texttt{i}(\tilde{x}_\texttt{i})$. Then, the combiner $\textbf{C}$ projects the output $\tilde{\textbf{Y}} = [\tilde{y}_\texttt{0}, \tilde{y}_\texttt{1}, ... \tilde{y}_{\texttt{E}\texttt{-}\texttt{1}}]$ to the token space $\textbf{Y} = \textbf{C}^T\tilde{\textbf{Y}}$.

\subsection{\sysname Block}
Conventional MoE design employs high rank experts in their MLP blocks that directly learn to handle different inputs.
Therefore, these experts are parameter-heavy and require expensive pretraining. 
The \sysname approach relies on adding (to an original base model denoted as $\theta^*$) experts that use a Soft MoE architecture, while simultaneously avoiding significantly increasing the parameter count by soft-mixing many zero-initialized {\it low-rank} experts.
Intuitively, the original base model $\theta^*$ serves as a foundational backbone, and the additional low-rank experts serve as ``specialists'' that gather additional specialized knowledge and handle different use cases. 

The base model $\theta^*$ can be initialized with either pretrained (raw), multitask-tuned, or instruction-tuned checkpoints. Using a raw checkpoint provides a more general backbone, while a multitask-tuned checkpoint provides a backbone focused on a required skill-set of the involved tasks -- we consider the decision of whether to use one or the other as a backbone to be application-dependent.
Our choice for Soft MoE \cite{softmoe} to instantiate the \sysname block follows from the desirable properties this architecture exhibits: fully differentiable, with no token dropping, and no expert balance issues.

The right part of Fig \ref{fig:framework} presents a \sysname block. \sysname operates on linear layers for the maximum flexibility and compatibility. We denote $\texttt{W}^*$ ($\texttt{W}^* \in \texttt{R}^{\texttt{d}_1 \times \texttt{d}_2}$) as the weight matrix of a linear layer in the base model $\theta^*$ and $\textbf{X} \in \texttt{R}^{\texttt{N} \times \texttt{d}_1}$ as the input with $N$ tokens. Following \cite{softmoe}, we introduce the routing matrix $\Phi \in \texttt{R}^{\texttt{E} \times \texttt{d}_1}$ and compute the dispatcher $\textbf{D} \in \texttt{R}^{\texttt{E} \times \texttt{N}}$ and the combiner $\textbf{C} \in \texttt{R}^{\texttt{E} \times \texttt{N}}$  using Eq. \ref{eq:dispatcher} and \ref{eq:combiner} for the $\texttt{E}$ experts. 

\sysname adopts a LoRA-inspired approach for the expert blocks. We introduce trainable low-rank matrices $\texttt{W}_i^{\texttt{out}}$, $\texttt{W}_i^{\texttt{in}}$ for the $i$-th expert, producing the output $\tilde{y}_i$ as in Eq.~\ref{eq:expert_lora}.
\begin{align}
\tilde{y}_i =   \texttt{W}_i^{\texttt{out}}\texttt{W}_i^{\texttt{in}}(\textbf{D}\textbf{X})_{i, :}^T\label{eq:expert_lora}
\end{align}
Then, the output of the \sysname \textbf{Y} combines the outputs of each expert and the original linear outputs, as in Eq. \ref{eq:final}.
\begin{align}
\textbf{Y} = \textbf{X}\texttt{W}^* + \textbf{C}^T[\tilde{y}_\texttt{0}, \tilde{y}_\texttt{1}, ... \tilde{y}_{\texttt{E}\texttt{-}\texttt{1}}]\label{eq:final}
\end{align}

\subsection{Omni-\sysname}
By default, \sysname blocks take as inputs all the tokens, regardless of their modality (denoted by $\mathcal{S}$MoLA$_\texttt{MM}$ in the next section).
However, we note that various multimodal tasks may place a different emphasis on how different modalities are used. For example, image captioning relies more on the visual tokens, VQA tasks on text-heavy images and using upstream OCR focuses more on text, while natural-image VQA must rely on both the visual and text tokens.  

Inspired by \cite{beitv3}, \sysname can be seamlessly configured to only adapt tokens for selected modalities. We denote the \sysname blocks that only take visual tokens or text tokens as $\mathcal{S}$MoLA$_\texttt{V}$ or $\mathcal{S}$MoLA$_\texttt{T}$, respectively.
$\mathcal{S}$MoLA$_{\texttt{MM}}$ refers to the \sysname blocks that take both visual and text tokens.
As shown in Figure \ref{fig:framework}, Omni-\sysname (denoted by $\mathcal{S}$MoLA$_\texttt{O}$ in the next section) combines via sum the original backbone outputs with the outputs of $\mathcal{S}$MoLA$_{\texttt{MM}}$ and the concatenated outputs of $\mathcal{S}$MoLA$_\texttt{V}$ and $\mathcal{S}$MoLA$_\texttt{T}$.

\subsection{The Properties of Omni-\sysname}
\noindent\textbf{Parameter Efficiency and Time Complexity.} The integration of LoRA and Soft MoE results in a combination that achieves a substantial reduction in the number of parameters required for adaptation, compared to traditional MoE~\cite{Lepikhin2020GShardSG}.
The low-rank matrices introduced by LoRA are of significantly lower dimensionality than the full-rank feedforward matrices, ensuring that the parameter increase is minimal (and controlable via the rank hyperparameter).
Not only does this lead to a leaner model, but it also reduces memory requirements, making it feasible to increase the number of experts to enhance performance.

Moreover, the inference cost of applying Omni-\sysname is negligible. Let $\texttt{d}_{\texttt{max}}$ denote $\texttt{max} \{\texttt{d}_1, \texttt{d}_2\}$ and $\texttt{r}$ denote the rank per expert, the time complexity of  \sysname blocks per-layer is $O(\texttt{E}\texttt{N}\texttt{d}_{max} + \texttt{E}(\texttt{d}_1 + \texttt{d}_2)\texttt{r})$. For one single layer, it increases the cost from $O(\texttt{N}\texttt{d}_1\texttt{d}_2)$ to $O(\texttt{N}\texttt{d}_1\texttt{d}_2 + \texttt{E}\texttt{N}\texttt{d}_{max} + \texttt{E}(\texttt{d}_1 + \texttt{d}_2)\texttt{r})$. The number of expert $\texttt{E}$ is always much smaller than $\texttt{min} \{\texttt{d}_1, \texttt{d}_2\}$, while the rank $\texttt{r}$ (typically a small integer like $4$) is much smaller than the input tokens length, especially for multimodal settings where a single high resolution image may easily be responsible for thousands of visual tokens. 

\noindent\textbf{Alternative Scaling Dimension.}
Traditional scaling methods in neural networks often involve increasing the size of the model, either by adding more layers or increasing the dimensionality of the existing layers. The proposed method, on the other hand, introduces an alternative scaling dimension. By leveraging sparse activation and parameter-efficient adaptation, the proposed method achieves scaling through increasing the number of the low-rank experts, which in turn does not result in a severe increase of total model parameter size.

\noindent \textbf{Extensibility for Future Growth.} The design of the proposed method inherently supports extensibility, accommodating future growth and adaptations with ease. As the requirements of a task evolve, additional low-rank specialist modules can be seamlessly integrated into the architecture, enhancing the model’s capability without necessitating a complete overhaul. This stands in stark contrast to traditional scaling methods, which often require predefined dimensions and layer numbers, limiting the model’s adaptability to changing scenarios.

\section{Experiments}
\subsection{Experimental setups}

\begin{table*}[ht!]
\centering
\resizebox{0.9\linewidth}{!}{%
\begin{tabular}{llccccccccccccccc}
\toprule
 & & COCO           && NoCaps$^\dag$     && VQAv2     && OKVQA && \multicolumn{2}{c}{A-OKVQA$^\dag$} && Sci-QA$^\dag$ && \multicolumn{2}{c}{TallyQA$^\dag$} \\
 \cmidrule{3-3} \cmidrule{5-5} \cmidrule{7-7}  \cmidrule{9-9}  \cmidrule{11-12} \cmidrule{14-14} \cmidrule{16-17}
& Model & Kar.-test && val        && test-dev  && val  && DA & MC && test && simple & complex \\
\midrule
\parbox[t]{2mm}{\multirow{6}{*}{\rotatebox[origin=c]{90}{Specialist}}} & GIT2~\cite{wang2022git}            & 145.0          && 126.9 && 81.7  && -        && - & -  &&    && - & - \\
& BEiT-3~\cite{beitv3}                 & 147.6          &&  -    && 84.2  && -        && - & -  &&    &&- & -\\
& PaLM-E~\cite{driess2023palme}       & 138.7          &&  -    && 80.0  && 66.1     && - & -  &&    &&- & -\\
& InstructBLIP ~\cite{dai2023instructblip}                     &   -            && 123.1 &&  -    && 62.1     && 62.1 & 73.4&& 90.7 &&  - & - \\
& PaLI-X~\cite{chen2023palix}                 & 149.2          && 126.3 && 86.0  && 66.1     && - & -  &&    && 86.0 &  75.6\\
& CogVLM \cite{cogvlm}                                   & 148.7          && 128.3 && 84.7  && 64.7     && - & -  &&  92.7 && - &  -\\
\midrule
\parbox[t]{2mm}{\multirow{7}{*}{\rotatebox[origin=c]{90}{Generalist}}}  & Unified-IO   \cite{lu2022unified}                & 122.3       && 100.0 && 77.9  &&  54.0     &&  45.2 & -  &&  -  && - &  -\\
 & Qwen-VL \cite{qwenvl}                  & -  && 121.4 && 79.5  && 58.6     && - & - && 67.1 && - & - \\

& CogVLM   \cite{cogvlm}                                 & 147.0          && \underline{126.2} && 83.4  && 58.9     && - & -  &&  -  && - &  -\\
 & \pali{3}{FT}                     & 144.4 && 120.3 && 82.5     && 56.2    && 59.0 & 78.7  && 55.2 && 80.4 & 65.4 \\
& \sysversion{48}{3}{FT}                     & 146.5 && 120.3 && 83.6  && 58.2    && 59.8 & 79.3 && 55.8 && 81.8 &  65.1  \\
 & \pali{X}{FT}                    & 148.7 && 125.6 && 84.4     && 60.7    && 63.9 & 84.0  && 67.2 && \underline{83.8} & \underline{71.8} \\
& \sysversion{48}{X}{FT}                     & \underline{\textbf{149.8}} && 126.1 && \underline{85.0} && \underline{62.4}     && \underline{\textbf{65.3}} & \underline{\textbf{84.1}} && \underline{67.8} &&  83.3 & 70.7  \\
\bottomrule
\end{tabular}}
\caption{Results on natural image captioning and question answering including COCO Captions (Karpathy split), NoCaps, VQAv2, OKVQA, A-OKVQA, ScienceQA and TallyQA test split with end-to-end modeling without OCR pipeline input. Bold and underlined numbers highlight best performance and best generalist performance, respectively. $^\dag$ denotes that there are no training examples from these datasets during training (i.e. out-domain). The numbers in bracket denote the further per-task LoRA tuned performances. We use the same \sysversion{48}{X}{FT} and \sysversion{48}{3}{FT} to handle inferences in Table 1 and Table 2. 
}
\label{table:natural_image}
\end{table*}

\begin{table*}[ht!]
\centering
\resizebox{0.8\linewidth}{!}{%
\begin{tabular}{llc@{\hspace{0.2cm}}c@{\hspace{0.2cm}}c@{\hspace{0.2cm}}c@{\hspace{0.2cm}}c@{\hspace{0.2cm}}c@{\hspace{0.2cm}}c@{\hspace{0.2cm}}c@{\hspace{0.2cm}}c@{\hspace{0.2cm}}c@{\hspace{0.2cm}}}
\toprule
& & Text & VizWiz & Text & VizWiz & ST & OCR & Info & Doc & \multirow{2}{*}{AI2D} & Chart \\ 
& Model & Caps & Cap & VQA & VQA & VQA & VQA & VQA & VQA & & VQA \\ \midrule
 &  & val & test & test & test-dev  & test & test & test & test &test & test \\ \midrule
 \multicolumn{4}{l}{\textit{\textbf{without} OCR pipeline input}} &&&&&&&\\\midrule
& Specialist SOTA & 158.8 \cite{chen2023pali5b} &  122.7 \cite{chen2023palix} & 79.5\cite{chen2023pali5b} & 76.4\cite{cogvlm} & 84.1\cite{chen2023pali5b} & 76.7\cite{chen2023pali5b} & 57.8\cite{chen2023pali5b}$^\ddag$ &  87.6\cite{chen2023pali5b}$^\ddag$ & 81.2\cite{chen2023palix} & 70.9\cite{chen2023palix}  \\\hdashline
\parbox[t]{2mm}{\multirow{6}{*}{\rotatebox[origin=c]{90}{Generalist}}} & Unified-IO \cite{lu2022unified} &  - & - & - &  57.4 & - &  - & - & - & -  & -  \\
& Qwen-VL \cite{qwenvl} & -  & - & 63.8 &  - & - &  \underline{75.7} & - & 65.1 & 62.3 & 65.7  \\
& CogVLM \cite{cogvlm} & 151.3 & - & 68.1 & - & - &  74.1 & - & - & - & - \\
& mPLUG-DocOwl \cite{mplug2} & 111.9 & - & 52.6 & - & - & - & 38.2 & 62.2 & - & 57.4 \\\cdashline{2-12}
 & \sysversion{48}{3}{FT}  & \underline{156.7} & 119.8 & -$^*$ & 70.4 & \underline{83.8} & 72.8 & \underline{52.4} & \underline{84.5} & 75.6 & 68.9 \\
 & \sysversion{48}{X}{FT}  & 144.6 & \underline{120.3} & 70.5 & \underline{71.7} & 78.9 &  71.6 & 49.2 & 80.1 & \underline{\textbf{81.4}} &  \underline{\textbf{71.3 }}\\
\midrule\midrule
\multicolumn{4}{l}{\textit{\textbf{with} OCR pipeline input}} &&&&&&&\\\midrule
& Specialist SOTA & 161.0 \cite{chen2023pali5b} & 125.7 \cite{chen2023palix}& 80.8 \cite{chen2023palix} & 76.4\cite{cogvlm} & 85.7\cite{chen2023pali5b} & 77.8\cite{chen2023pali5b} & 62.4\cite{chen2023pali5b} & 88.6\cite{chen2023pali5b} & 81.4\cite{chen2023palix} & 72.3\cite{chen2023palix}  \\ \hdashline
 & \sysversion{48}{3}{FT} & 159.3 & 120.4 & -$^*$ & 71.0 &  85.9 & 73.9 & 57.3 & 87.4 & 75.5 & 68.9 \\
 & \sysversion{48}{X}{FT}  & 154.7 &  124.6 &  \textbf{81.1} &73.8 & \textbf{86.0} &  74.9 &  \textbf{65.6} & \textbf{90.6} & \textbf{81.4} &  \textbf{73.8} \\
\bottomrule
\end{tabular}}
\caption{Results on benchmarks more focused on text understanding capabilities. Bold and underlined numbers highlight SOTA performance and SOTA generalist performance, respectively. $^\ddag$ marks specialist results with a higher resolution of 1064 where \sysname used 812.  We use the same \sysversion{48}{X}{FT} and \sysversion{48}{3}{FT} to handle inferences with and without OCR pipeline input in Table 1 and Table 2. $^*$Results are missing because test server is not available.
}
\label{table:scene_text_sota}
\end{table*}

\paragraph{Training Mixtures.} We considers three mixtures: \\
\begin{itemize}
  \setlength\itemsep{-0.8em}
  \item{\textit{Image Captioning mixture}}: COCO captions\footnote{In keeping with the multilingual nature of PaLI models, here we used a variant of the original English-only COCO captions that included translated captions for an additional 35 languages.} \cite{cocokarp} , Textcaps \cite{textcaps}, VizWiz-Cap \cite{vizwizcap}.\\
  \item{\textit{VQA mixture}}: VQAv2\footnote{Included translated questions for an additional 13 languages.} \cite{vqav2}, OK-VQA \cite{marino2019okvqa}, VizWiz-VQA \cite{gurari2018vizwiz}, ST-VQA \cite{stvqa}, TextVQA \cite{textvqa}, OCRVQA \cite{ocrvqa}, InfoVQA \cite{mathew2022infographicvqa}, DocVQA \cite{mathew2021docvqa}, ChartQA \cite{masry2022chartqa}, AI2D \cite{ai2d}. \\
  \item{\textit{Full mixture}}: combines the Image Captioning mixture and the VQA mixture.\\
\end{itemize}

By default, we use the full mixture in our experiments to simulate the scenario of mixing a wide variety of different tasks.  The only exception is Sec. \ref{sec:ablation:mixture}, where we measure the effect of using more focused mixtures.

\paragraph{Task Prompts.} We do not use benchmark specific prompts in order to achieve better versatility of the generalist models. Following \cite{chen2023palix} and \cite{chen2023pali5b}, we use \texttt{Generate} \texttt{the} \texttt{alt\_text} \texttt{in} \texttt{\{lang\}} \texttt{at} \texttt{0:} as the captioning prompt and \texttt{Answer} \texttt{in} \texttt{en:} \texttt{\{question\}} as the VQA prompt. 
\paragraph{Base Models.} We build \sysname models on top of two variants of PaLI models: PaLI-X \cite{chen2023palix} and PaLI-3 \cite{chen2023pali5b}.  PaLI models use contrastively pretrained ViT modules as the visual encoder to produce visual embeddings for input images; these visual embeddings are then concatenated with text embeddings and passed to the encoder-decoder backbone.
PaLI-X is a large-scale multimodal model that contains around 55B parameters.  We only experimented with using the full-mixture in PaLI-X based experiments, where we adopted a resolution of 672.
PaLI-3 is a more nimble variant.  It is still highly performant with just around 5B parameters, achieving SOTA results on a broad range of image captioning and VQA tasks that require text understanding capabilities from images.  For PaLI-3 based experiments, we use a resolution of 812 for the full mixture and the image captioning mixture, and 1064 
for the VQA mixture.

\begin{table*}[ht!]
\centering
\resizebox{0.9\linewidth}{!}{%
\begin{tabular}{lc@{\hspace{0.2cm}}c@{\hspace{0.2cm}}c@{\hspace{0.2cm}}c@{\hspace{0.2cm}}c@{\hspace{0.2cm}}c@{\hspace{0.2cm}}c@{\hspace{0.2cm}}c@{\hspace{0.2cm}}c@{\hspace{0.2cm}}c@{\hspace{0.2cm}}c@{\hspace{0.2cm}}c@{\hspace{0.2cm}}c@{\hspace{0.2cm}}c@{\hspace{0.2cm}}}
\toprule
        & COCO  & Text & VizWiz   & VQA & OK & Text & VizWiz & ST & OCR & Info & Doc & \multirow{2}{*}{AI2D} & Chart & Avg. \\ 
Model   & Cap   & Cap & Cap & v2 &  VQA & VQA & VQA & VQA & VQA & VQA& VQA & & VQA & $\delta$ \\ \midrule 
  & K.test   & val & test & test-dev & val & val$^*$ & test-dev & test & test & test& test &test & test &  \\ \midrule
\multicolumn{8}{l}{\textit{\textbf{with} OCR pipeline input, except for COCO Cap, VQAv2, OKVQA}} &&&&&\\\midrule
PaLI-3 Specialist & 145.9 & 161.0 & 120.3 & 85.0 & 60.1 & 78.3
&  72.2 & 85.7 & 77.8 & 62.4$^\ddag$ & 88.6$^\ddag$ & 75.2 & 69.5 & 0.00 \\\hdashline
\sysversion{48}{3}{RAW} & 144.4 & 159.1 & 118.7 &  82.6 & 56.2 & 79.1 
&  70.6 & 85.5 & 73.3 & 55.1 & 86.6 & 73.8 & 67.6 & -2.26 \\\hdashline
PaLI-3$_\texttt{FT}$ & 146.2 & \underline{\textbf{161.0}} & 121.1 & 82.5 & 56.4 & 78.7 
& 69.9 & 84.9 & 72.7 & 54.3 & 85.9 & 72.8 & 65.8 & -2.31 \\
\sysversion[MM]{96}{3}{FT} & 145.7 & 159.3 & \underline{\textbf{121.4}} & 83.4 & 56.7 & 80.0
& \underline{71.5} & 85.6 & 73.6 & 56.7 & 87.3 & 75.2 & \underline{69.2} & -1.26\\
\sysversion{48}{3}{FT}& \underline{\textbf{146.5}} & 159.3 & 120.4 & \underline{83.6} & \underline{58.2} & \underline{\textbf{80.1}} 
& 71.0 & \underline{\textbf{85.9}} & \underline{73.9} & \underline{57.3} & \underline{87.4} & \underline{\textbf{75.5}} & 68.9 & \underline{-1.07}  \\\midrule\midrule
  & K.test   & val & test & test-dev & val & test & test-dev & test & test & test& test &test & test &  \\ \midrule
PaLI-X Specialist & 149.2 & 159.6 & 125.7 & 86.0$^\ddag$ & 66.1$^\ddag$ & 80.8$^\ddag$ & 74.6$^\ddag$ & 84.5$^\ddag$ & 77.3$^\ddag$ & 54.8$^\ddag$ & 86.8$^\ddag$ & 81.4$^\ddag$ & 72.3$^\ddag$ & 0.00 \\\hdashline
PaLI-X$_\texttt{LoRA}$ & 147.3 & \underline{159.3} & 125.1 & 83.5 & 57.4 & 78.9 & 69.6 & 84.8 & 72.3 & 61.4 & 88.3 & 78.8 & 70.9 & {-1.65}\\
\sysversion{48}{X}{LoRA} & 148.6 & 158.8 & 125.2 & 84.7 & 60.8 & 80.3 & 73.1 & 85.2 & 74.2 & 64.8 & 90.1 & 80.2 & 73.0 & -0.01 \\ \hdashline
PaLI-X$_\texttt{FT}$ & 148.7 & 157.0 & \underline{125.3} & 84.4 & 60.7 & 79.6 & 72.2 & 84.7 & 73.5 & 62.4 & 88.2 & 80.7 & 70.2 & -0.88 \\
\sysversion{48}{X}{FT} & \underline{\textbf{149.8}} & 154.7 & 124.6 & \underline{85.0} & \underline{62.4} & \underline{\textbf{81.1}} & \underline{73.8} & \underline{\textbf{86.0}} & \underline{74.9} & \underline{\textbf{65.6}} & \underline{\textbf{90.6}} & \underline{\textbf{81.4}} & \underline{\textbf{73.8}} & \underline{\textbf{+0.38}} \\
\bottomrule
\end{tabular}}
\caption{Ablation results on image captioning and question answering benchmarks. Bold and underlined numbers highlight best performance and best generalist performance, respectively. $^\ddag$ denotes the specialist results with a higher resolution of 1064 for PaLI-3 and 756 resolution for PaLI-X, where we uses 812 for PaLI-3 series and 672 for PaLI-X series. $^*$We use val split as TextVQA test server is broken.
}
\label{table:scene_text_ablation}
\end{table*}

\paragraph{Notation and implementation.} We use \sysversion[Y]{E}{3$|$X}{{RAW|LoRA|FT}} to denote the config choices for \sysname: 
\begin{itemize}
    \item \texttt{E} denotes the number of experts for each individual modality and for multimodal experts.
    \item \texttt{Y} denotes the \sysname's modality configuration: \texttt{MM} or \texttt{O}.
    \item base model: PaLI-3 vs PaLI-X
    \item \sysname's initial checkpoint can be either the {\footnotesize \texttt{RAW}} checkpoint of the base model, the base model tuned using {\footnotesize \texttt{LoRA}} on a given training mixture, or full-model fine-tuned ({\footnotesize \texttt{FT}}) using the training mixture. We use a rank of 128 for LoRA tuning on all linear layers on the PaLI encoder.\footnote{LoRA with rank 512  did not achieve better overall performance.}
\end{itemize}
For simplicity, we assign the same number of experts to each \sysname block and use a rank of 4 per expert. \sysname is applied on all the linear layers in the attention and MLP modules in PaLI encoder blocks.  For example, \sysversion{48}{X}{FT} with full-mixture denotes starting with PaLI-X finetuned on the full-mixture, and then \sysname-tuned on the same mixture using 48 visual-token experts, 48 text-token experts, and 48 multimodal-token experts. 
\paragraph{Checkpoint selection.} We monitor the scores on the validation splits\footnote{We use the Pix2Struct validation split for AI2D.} every 500 iterations with at most 1,024 examples for each task and select the checkpoint with maximum average validation scores.

\subsection{Main Results}
\label{exp:main}

In this section, we present our main experimental results using the full mixture.
Recall that the full mixture contains both image captioning and VQA tasks.  We report \sysname results on the natural image tasks (as well as ``out-domain'' tasks not included in the training mixture) in Table \ref{table:natural_image}, and results on tasks that focus on understanding texts in images in Table \ref{table:scene_text_sota}.  While results are split into these two tables for easier consumption, they are from the same \sysname-based generalist models trained on one single mixture.  

First, note that the generalist \pali{X}{FT} (PaLI-X finetuned on the full-mixture) under-performs its specialist counterparts (PaLI-X finetuned for each task individually) on all the benchmark datasets shown in Table \ref{table:natural_image}.
Applying \sysname over \pali{X}{FT} outperformed the base generalist model across the board. It effectively shortened the gap to specialist performances, and notably introduced a new SOTA CIDEr score of 149.8 on COCO captioning, outperforming all the specialist models for that task.

\begin{table*}[ht!]
\centering
\resizebox{0.7\linewidth}{!}{%
\begin{tabular}{lc@{\hspace{0.2cm}}c@{\hspace{0.2cm}}c@{\hspace{0.2cm}}c@{\hspace{0.2cm}}c@{\hspace{0.2cm}}c@{\hspace{0.2cm}}c@{\hspace{0.2cm}}c@{\hspace{0.2cm}}c@{\hspace{0.2cm}}c@{\hspace{0.2cm}}c@{\hspace{0.2cm}}}
\toprule
     & \multirow{2}{*}{VQAv2} & OK & Text & VizWiz & ST & OCR & Info & Doc & \multirow{2}{*}{AI2D} & Chart & Avg. \\ 
Model &   & VQA & VQA & VQA & VQA & VQA & VQA & VQA & & VQA & $\delta$\\ \midrule
  &  test-dev & val & test & test-dev & test & test & test & test & test & test &  \\ \midrule
\multicolumn{4}{l}{\textit{\textbf{without} OCR pipeline input}} &&&&&&&\\\midrule
PaLI-3 Specialist & 85.0 &  60.1 & 79.5 & 71.9  & 84.1 & 76.7 & 57.8 &  87.6 & 75.2 & 70.0 & 0.0 \\\hdashline
 PaLI-3$_\texttt{FT}$  & 82.1 & \underline{57.9} & 79.8 & 69.2 & 84.0 &  72.5 & 55.9 & 87.6 & 74.2 & 68.0 & -1.53 \\
 \sysversion{48}{3}{FT} & \underline{83.4} & 57.7 & \underline{\textbf{80.0}} & \underline{70.8} & 84.0 &  \underline{73.4} & \underline{57.3} & \underline{\textbf{87.8}} & \underline{\textbf{75.9}} & \underline{\textbf{70.1}} & \underline{-0.46} \\
\midrule\midrule
\multicolumn{4}{l}{\textit{\textbf{with} OCR pipeline input}} &&&&&&&\\\midrule
PaLI-3 Specialist & - & - & 80.8  & 72.2 & 85.7 & 77.8 & 62.4 & 88.6 & 75.2 & 69.5 & \\ \hdashline
 PaLI-3$_\texttt{FT}$  & - & - &  81.7 & 70.0 & 85.5 & 73.6 & 59.8 & 88.8 & 74.7 &  67.3 & \\
 \sysversion{48}{3}{FT} & - & - &  \underline{\textbf{82.2}} & \underline{72.0} & \underline{\textbf{85.8}} &  \underline{74.6} & \underline{61.1} & \underline{\textbf{89.3}} & \underline{\textbf{76.0}} & \underline{ \textbf{70.4}} & \\
\bottomrule
\end{tabular}}
\caption{Generalist results using the VQA mixture. Bold numbers highlight the results outperforming single specialized PaLI-3 baselines, and underlined numbers presents the results outperform multi-task fine-tuned baselines.}
\label{table:scene_text_ablation_vqa}
\end{table*}

It is important to note that Table \ref{table:natural_image} presents results for both  ``in-domain'' tasks that are included in the training mixture (COCO captioning, VQAv2, and OKVQA), as well as ``out-domain'' tasks (those marked with $^\dag$).  The in-domain tasks simulate usecases where we are interested in serving one single model for a set of known tasks.  The out-domain tasks simulate usecases where we want to apply a generalist model to unseen tasks in a zero-shot setting.  The trend we noted above holds for both cases: 
\sysversion{48}{X}{FT} outperforms base model \pali{X}{FT}  for both in-domain and out-domain tasks on average.
Overall, \sysversion{48}{X}{FT} achieves new SoTA generalist results for all except NoCaps and TallyQA,
and furthermore beating fine-tuned specialist models for COCO (in-domain) and A-OKVQA (out-domain). 
While \pali{3}{FT} based models overall under-performs \pali{X}{FT} based models on this set of tasks, \sysname nonetheless improves the base model performance consistently, demonstrating the effectiveness of this technique for both large- and small-scale models.

Table \ref{table:scene_text_sota} presents  \sysname results on text-heavy tasks in two experimental setups: (a) relying solely on a model's text understanding capabilities from the raw pixels ({\em without} OCR input), and (b) including tokens extracted by an upstream OCR module as part of the text input ({\em with} OCR input).
In the with-OCR setting,
\sysversion{48}{X}{FT} shows remarkable results: with one single model, it outperforms specialist SoTA performance on 6 out of 10 datasets, yielding new SoTA performance for TextVQA, ST-VQA, InfoVQA, DocVQA, AI2D and ChartQA.  It also improves over the base model \pali{x}{FT} (see Section \ref{sec:ablation:base}). This indicates that \sysname is effective in enabling joint processing of information across different modalities: text situated in image, as well as text tokens extracted by the upstream OCR module. 
In the without-OCR setting, \sysversion{48}{3}{FT} is able to take advantage of PaLI-3's strong text understanding capability and achieves SOTA generalist score on TextCaps, 
ST-VQA, InfoVQA, and DocVQA.

\subsection{Ablation Studies}

\subsubsection{Different base models}
\label{sec:ablation:base}
In Section \ref{exp:main}, we see strong performance from \sysversion{48}{X}{FT}, starting from a strong checkpoint (full-model PaLI-X finetuned).
In this section, we examine the effect of switching to PaLI-X$_\texttt{LoRA}$, which is LoRA-tuned on the mixture and easier to obtain for large models.
As shown in Table \ref{table:scene_text_ablation}, compared to their corresponding base models, we find \sysname helps both PaLI-X$_\texttt{LoRA}$ and PaLI-X$_\texttt{FT}$ to achieve better overall results, obtaining +1.64 and +1.26 improvements on average, respectively.
While it is slightly weaker than \sysversion{48}{X}{FT}, which outperforms per-task fine-tuned specialist models by an average of 0.38 points, \sysversion{48}{X}{LoRA} still achieves competitive performance versus the specialist models (on average only a difference of 0.01 point).
It is worth noting that the \sysname design improves PaLI-X$_\texttt{FT}$ by +2.4 points on DocVQA, +3.2 points on InfoVQA, and +3.6 points on ChartQA, which all involve comprehending rich text and symbols in images. We note some performance drop on the TextCaps task, possibly due to overfitting and unambiguous intention for image captioning tasks when the same prompt is used for TextCaps and natural-image descriptions. 

Table \ref{table:scene_text_ablation} also shows other ablation results on using different starting checkpoints ($\theta^*$) for \sysname. 
Similar observation holds for the PaLI-3-based models.  For instance, applying \sysname to the raw checkpoint ($i.e.$ PaLI-3$_\texttt{RAW}$) achieves better overall score than full model fine-tuning baseline PaLI-3$_\texttt{FT}$,
and applying \sysname to PaLI-3$_\texttt{FT}$ brings it more competitive against PaLI-3 specialists, outperforming per-task finetuned baselines on 4 benchmarks.  One exception is InfoVQA  where the specialist uses a higher resolution.

\subsubsection{Effect of Using Multi-Modal Experts}
We validate the omni experts design by comparing the average performance of using 48 experts on each combination of modalities ($i.e.$ \sysversion{48}{3}{FT}) to using 96 experts on all tokens ($i.e.$ \sysversion[MM]{96}{3}{FT}).  These two variants introduce the same additional FLOPS during inference.
As shown in Table \ref{table:scene_text_ablation}, \sysversion{48}{3}{FT} has a slightly edge in terms of average performance.
This suggests that for the same amount of extra compute, there can be a slight advantage to allow modality-dependent  
\sysname blocks.

\subsubsection{Effect of Scaling Up the Expert Counts}
\begin{figure}[h]
    \centering
    \includegraphics[clip, trim=0cm 0.5cm 0cm 0.7cm, width=0.7\columnwidth]{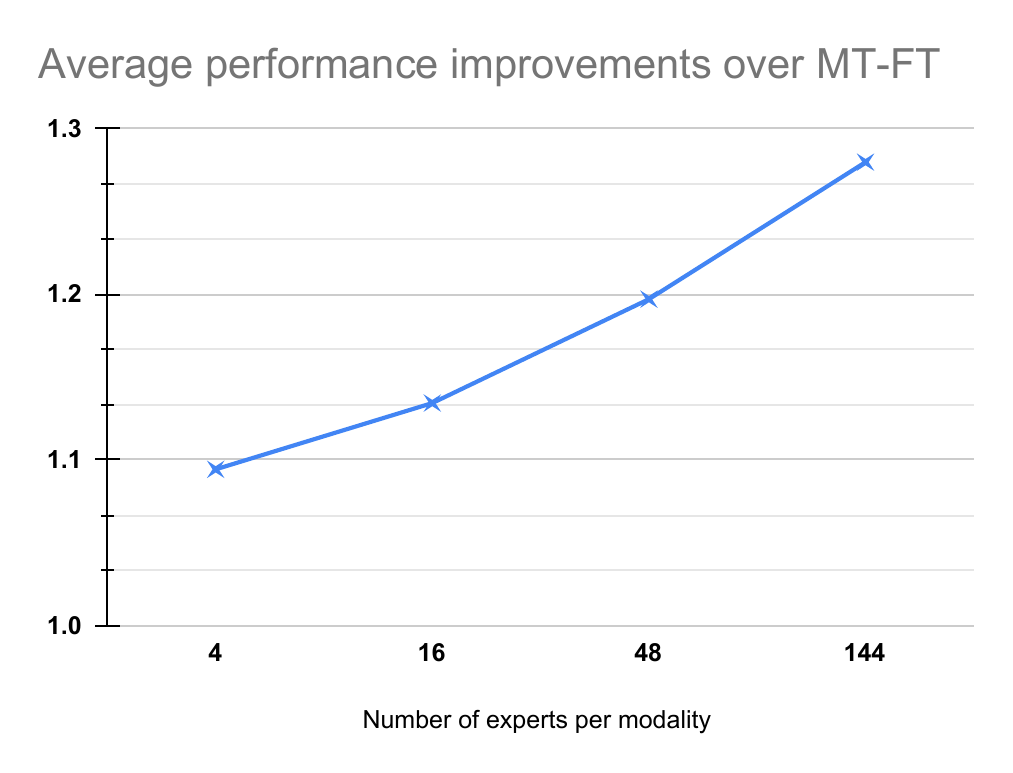}
    \caption{Average results of increasing number of experts.}
    \label{fig:scaling_experts}
\end{figure}
We study the effect of scaling up the expert counts using \sysversion{E}{3}{FT}. Figure \ref{fig:scaling_experts} plots the average improvements over  PaLI-3$_\texttt{FT}$ using 4, 16, 48, and 144 experts per modality. 
The scores are averaged across the tasks presented in Table \ref{table:scene_text_ablation} on validation splits\footnote{We use test split for AI2D as there are only 120 examples in val split.} except for InfoVQA. 
With only 4 experts per modality, \sysname already yields around +1.1 improvements over \pali{3}{FT}. Scaling up the experts counts further improves the performance: 16, 48 and 144 experts provide 1.14, 1.2, 1.27 average points gain. 

\subsubsection{Further LoRA tuning to Push SOTA}
We note there that the \sysversion{48}{X}{FT} generalist can be considered a strong foundational model.
With further per-task LoRA tuning (using a rank of 4), the \sysversion{48}{X}{FT} specialists achieve better results than the \sysversion{48}{X}{FT} generalist, yielding new SOTA results on 9 benchmark datasets: COCO caption, OKVQA, DocVQA, InfoVQA, AI2D, ChartQA, A-OKVQA, ScienceQA and TallyQA (Table \ref{tab:new_sota}). These new SOTA results indicate the extensibility of the  Omni-\sysname design.

\begin{table}[h]
\centering
\resizebox{0.8\linewidth}{!}{
\begin{tabular}{l|l|c|c|c}
\toprule
       Model                  &    Split     & PaLI-X & SOTA  &  Ours \\\midrule
COCO                     & K.test         & 149.2  & 149.2 \cite{chen2023palix} & \textbf{152.1} \\
VQA v2                     & test-dev         & 86.0  & \textbf{86.0} \cite{chen2023palix} & 85.7 \\
OKVQA                    & val            & 66.1   & 66.1 \cite{chen2023palix}  & \textbf{66.7}  \\
VizWiz-VQA                  & test-dev       & 74.6   & \textbf{76.4} \cite{cogvlm}  & 75.9  \\
OCRVQA                  & test     & 77.3   & \textbf{77.8} \cite{chen2023pali5b}  & 75.7  \\
DocVQA                   & test           & 86.8   & 88.6 \cite{chen2023pali5b}  & \textbf{90.8}  \\
InfoVQA                  & test           & 54.8   & 62.4 \cite{chen2023pali5b} & \textbf{66.2}  \\
AI2D                     & test           & 81.4   & 81.4 \cite{chen2023palix} & \textbf{82.5}  \\
ChartQA                  & test           & 72.3   & 72.3 \cite{chen2023palix} & \textbf{74.6}  \\
\multirow{2}{*}{A-OKVQA}  & DA (val)       & -     & 62.1 \cite{dai2023instructblip}   & \textbf{70.2}  \\
                         & MC (val)       & -      & 73.4 \cite{dai2023instructblip} & \textbf{88.2}  \\
ScienceQA                & test           & -      &92.7 \cite{cogvlm}  & \textbf{94.7}  \\
\multirow{2}{*}{TallyQA} & simple  & 86.0      &86.0 \cite{chen2023palix}  & \textbf{86.3}  \\
                         & complex & 75.6       &75.6 \cite{chen2023palix}  & \textbf{77.1} \\\bottomrule
\end{tabular}}
\caption{Further LoRA tuning  \sysversion{48}{X}{FT} }
\label{tab:new_sota}
\end{table}

\subsubsection{Inference Speed Comparison}
We compare the inference speed by measuring the number of processed examples
per second (eps) for the \pali{3}{FT} model and \sysversion{48}{3}{FT} with a resolution of 812 in batch mode (size 128) using beam decoding (beam size 4)
We use COCO caption as the evaluation task where the length of outputs are around 10 tokens on average. We sample 18 forward batches to compute the statistics. \pali{3}{FT} processed $31.29 \pm 0.63$ examples per second and \sysversion{48}{3}{FT} processed $30.85 \pm 0.70$ examples per second, yielding only 1.4\% slow-down when using 48 experts in each \sysname block, on all linear layers in the PaLI encoder.

\subsubsection{Effects of different training mixtures}
\label{sec:ablation:mixture}
We evaluate \sysname  on the VQA and captioning mixture with PaLI-3 in order to exam its effectiveness when all training tasks are under the same umbrella of either VQA or captioning. We adopt a resolution of 1064 for the VQA mixture and 812 for the captioning mixture, and finetune the PaLI-3 raw checkpoints on each mixture as baselines.

\noindent\textbf{VQA Mixture}  As shown in Table \ref{table:scene_text_ablation_vqa}, while  still underperforming the per-task fine-tuned specialist models by 0.46, \sysname improves over the PaLI-3$_\texttt{FT}$ baseline by +1.07 on average. In particular, it helps the base model on most of the tasks with and without OCR inputs except for a 0.2 performance drop on OK-VQA. The significant performance improvements over the PaLI-3$_\texttt{FT}$  baseline on InfoVQA and ChartQA persist as observed when training with the full mixture. Furthermore, 
it also helps the PaLI-3-based model to achieve a new SOTA result of 82.2 on TextVQA.\\

\noindent\textbf{Image Captioning Mixture}
Table \ref{table:scene-text-like} summarizes results of applying \sysname to PaLI-3$_{\texttt{LoRA}}$ and PaLI-3$_{\texttt{FT}}$ using the image captioning mixture.  In these experiments, we use a resolution of $812$. We observe similar trends as in the case of using the full mixture: PaLI-3$_{\texttt{FT}}$ outperforms PaLI-3$_{\texttt{LoRA}}$ on average, indicating the insufficiency of LoRA tuning on a wide range of tasks; and \sysname helps both the full model fine-tuned and LoRA baseline achieve better average performance. The 
\sysversion{48}{3}{FT} outperforms the per-task fine-tuned specialist models by 0.64 on average, and sets a new SOTA for a generalist image-captioning system.

\begin{table}[h]
\centering
\resizebox{\linewidth}{!}{%
\begin{tabular}{lcccccc}
\toprule
     &  \multirow{2}{*}{COCO}     & \multicolumn{2}{c}{TextCap} & \multicolumn{2}{c}{VizWizCap} & Avg. \\ 
Model & &   $ocr \times$ & $ocr \checkmark$  & $ocr \times$ & $ocr \checkmark$  &  $\delta$\\ \midrule

     & K. test  & {val} & {val} & test & test & \\ \midrule
PaLI-3 Specialist & 145.9   & 158.8 & 161.0 & 119.6 & 120.3 & 0.0 \\\hdashline
PaLI-3$_\texttt{LoRA}$ & 143.6 & 158.6 & 161.3 & 118.8 & 120.5  & -0.56 \\
\sysversion{48}{3}{LoRA} & 143.9 &  160.6 & \underline{\textbf{162.6}}	& 119.1 &120.8 & +0.28 \\\hdashline
PaLI-3$_\texttt{FT}$ &  145.0 & \underline{\textbf{159.9}} & 160.9 & 120.3 & \underline{\textbf{120.9}} &  +0.28 \\
\sysversion{48}{3}{FT}& \underline{\textbf{146.5}} & 159.5 & 161.7 & \underline{\textbf{120.5}} & 120.6  &  \underline{\textbf{+0.64}}
\\
\bottomrule
\end{tabular}}
\caption{Generalist results using the image captioning mixture. Bold and underlined numbers highlight best performance and best generalist performance, respectively.
}
\label{table:scene-text-like}
\end{table}

\section{Conclusion}
In this work, we present Omni-\sysname, a multimodal architecture that mixes many multi-modal experts efficiently and achieves both high specialist and generalist performance.
In contrast to previous models for which we see performance degradation on average when training the models on a wide range of tasks, we show that the \sysname low-rank experts are able to model different skills and task, and overall improve the performance of a generalist model.
This finding indicates that simple LMM fine-tuning is suboptimal for handling a wide range of tasks, and that pairing the act of fine-tuning with specifically-designed architecture changes leads to better performing models.

{
    \small
    \bibliographystyle{ieeenat_fullname}
    \bibliography{main}
}

\appendix

\section{Scaling up \sysname vs LoRA}
\subsection{Effect of scaling up rank for LoRA}
Recall that in Section \ref{sec:exp:expertcounts}, we studied the effect of scaling up the expert counts for \sysversion{E}{3}{FT}.  Specifically,  with $\texttt{E=}$ 4, 16, 48, and 144 experts per modality, \sysversion{E}{3}{FT} models achieve 1.1, 1.14, 1.2, 1.27 improvements over the \pali{3}{FT} baseline, where the metrics are averaged across the tasks presented in Table 3 in the main paper on the validation splits\footnote{We use test split for AI2D as there are only 120 examples in val split.} except for InfoVQA.   Can we achieve similar performance improvement by using a higher rank of LoRA on top of \pali{3}{FT}?  

We experiment with LoRA tuning on \pali{3}{FT} with rank $\texttt{r}$ of 32, 128, 384, 1536.  Note that LoRA tuning with rank $\texttt{r}$ has the same extra model parameters as \sysversion[MM]{r/4}{3}{FT}, and  shares the same extra compute as \sysversion{r/8}{3}{FT}. These LoRA tuning runs brought 0.58, 0.86, 0.79 and 0.59 gain over the \pali{3}{FT} baseline. At lower ranks ($\texttt{r} = 32, 128$), LoRA tuning led to smaller gains than \sysversion{r/8}{3}{FT}, and further scaling up the rank resulted in worse performance in the case of LoRA tuning.

\subsection{Effective rank for LoRA}

\begin{figure}[h]
  \centering
  \includegraphics[clip, trim=0cm 6cm 0cm 0cm, width=\columnwidth]{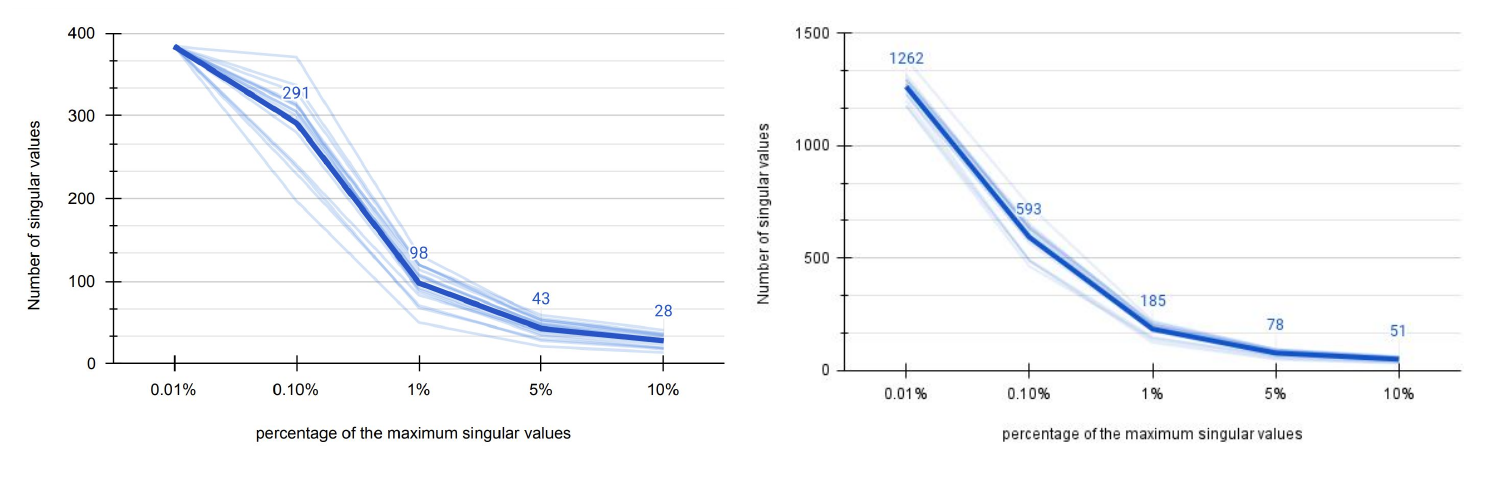}
  \caption{Number of singular values that are greater than different thresholds for LoRA weights. The left is for LoRA with rank 384 and the right is for LoRA with rank 1536.}
  \label{fig:rank_viz}
\end{figure}

To help understand why higher ranks do not improve LoRA tuning,
we investigate how many ranks have the potential to significantly impact the outputs. 
Specifically, we consider LoRA with rank 384 and 1536 applied to \pali{3}{FT}.
We compute the singular values for the combined weights of $\texttt{W}^{\texttt{out}}\texttt{W}^{\texttt{in}}$ for  the output layer of each attention module in the encoder blocks, and present statistics over singular values in these two settings in  Figure \ref{fig:rank_viz}.  In particular, we plot the number of singular values that are greater than 0.01\%, 0.1\%, 1\%, 5\%, 10\% of the largest singular value -- the light colored lines plot this for each encoder block and the bold line plots the counts averaged across all encoder blocks. For LoRA with rank 1536, there are only about 78 (5\%) and 185 (12\%) singular values greater than 5\% and 1\% of the largest singular value, the remaining dimensions would make little contribution to the output with normalized inputs.

\section{Visualization on the routing matrix $\Phi$}

\begin{figure}
  \centering
  \includegraphics[clip, trim=0cm 5cm 0cm 0cm, width=1.1\columnwidth]{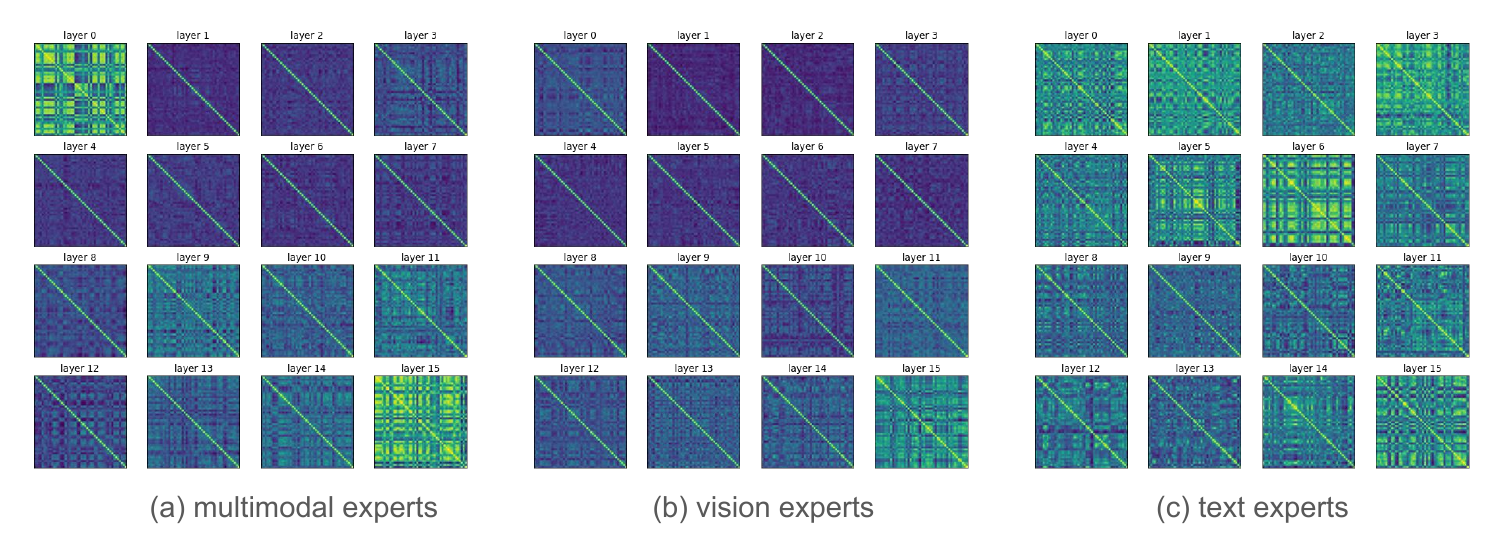}
  \caption{Heat maps for $\Phi \Phi ^T$ for different modalities in \sysversion{48}{3}{FT}}
  \label{fig:pali5b_viz}
\end{figure}

We present a heat map visualization of $\Phi \Phi ^T$ in Figure \ref{fig:pali5b_viz}. If $\Phi \Phi ^T$ closely resembles an identity matrix, the routing matrix  $\Phi$ is more likely to route the input to different experts. We have the following observations: (a)   $\Phi \Phi ^T$ for text experts are less similar to the identity matrix, indicating less need for specialization. One possible explanation is that input texts in our tasks tend to be easy to process --- we trained on VQA tasks with short questions or image captioning tasks using identical prompt. (b)  $\Phi \Phi ^T$ for early layers in vision and multimodal experts are much closer to the identity matrix, indicating the need for different experts to handle more diverse shallow representations.

\section{Pseudo Code}
\begin{figure}[h]
    \centering
    \includegraphics[clip, trim=0cm 0cm 0cm 0cm, width=\columnwidth]{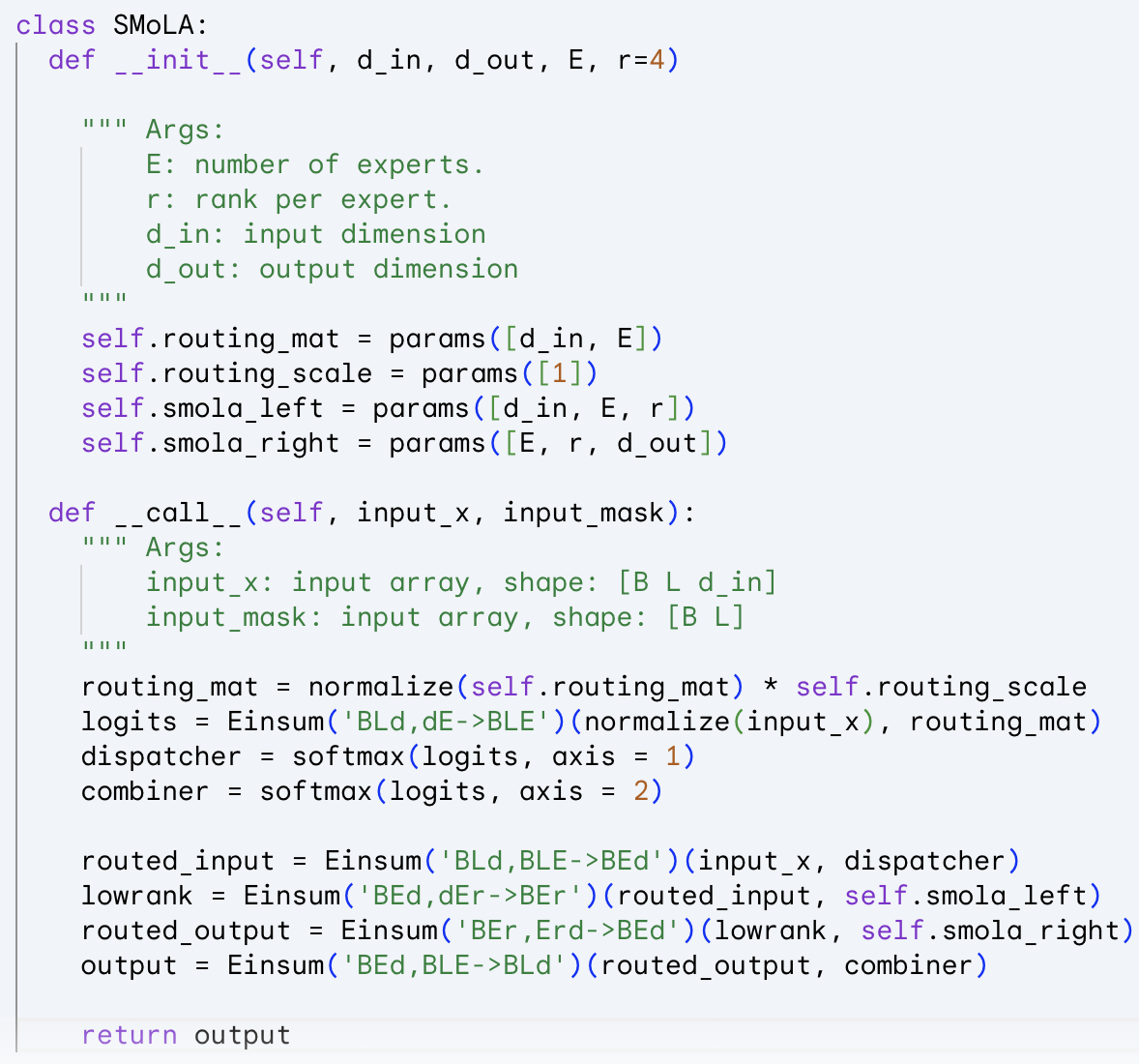}
    \caption{Pseudo Code for a SMoLA block}
    \label{fig:code}
\end{figure}

While we largely inherit the Soft MoE design, we provide the Pseudo Code for running SMoLA blocks in Figure \ref{fig:code}.

\end{document}